\documentclass[runningheads]{llncs}
\usepackage{comment}
\usepackage[T1]{fontenc}
\usepackage{dingbat}
\usepackage{todonotes}
\usepackage{hyperref}
\usepackage{graphicx}
\usepackage{multirow}
\begin{document}

\title{OntoAligner: A Comprehensive Modular and Robust Python Toolkit for Ontology Alignment}
\titlerunning{OntoAligner}
\author{Hamed Babaei Giglou\inst{1}\orcidID{0000-0003-3758-1454} 
       \and Jennifer D'Souza\inst{1}\orcidID{0000-0002-6616-9509}
       \and Oliver Karras\inst{1}\orcidID{0000-0001-5336-6899} 
       \and S\"{o}ren Auer\inst{1,2}\orcidID{0000-0002-0698-2864}
       }
\institute{TIB - Leibniz Information Centre for Science and Technology, Hannover, Germany \and
L3S Research Center, Leibniz University of Hannover, Germany
\\ 
         \email{\{hamed.babaei, jennifer.dsouza, oliver.karras, auer\}@tib.eu}}

\authorrunning{Babaei Giglou. H et al.}
%
%
%
\maketitle
\begin{abstract}
Ontology Alignment (OA) is fundamental for achieving semantic interoperability across diverse knowledge systems. We present OntoAligner, a comprehensive, modular, and robust Python toolkit for ontology alignment, designed to address current limitations with existing tools faced by practitioners. Existing tools are limited in scalability, modularity, and ease of integration with recent AI advances. OntoAligner provides a flexible architecture integrating existing lightweight OA techniques such as fuzzy matching but goes beyond by supporting contemporary methods with retrieval-augmented generation and large language models for OA. The framework prioritizes extensibility, enabling researchers to integrate custom alignment algorithms and datasets. This paper details the design principles, architecture, and implementation of the OntoAligner, demonstrating its utility through benchmarks on standard OA tasks. Our evaluation highlights OntoAligner’s ability to handle large-scale ontologies efficiently with few lines of code while delivering high alignment quality. By making OntoAligner open-source, we aim to provide a resource that fosters innovation and collaboration within the OA community, empowering researchers and practitioners with a toolkit for reproducible OA research and real-world applications.

\end{abstract}

\begin{itemize}
    \item[] \textbf{Resource type}: Software
    \item[] \textbf{License}: Apache License 2.0
    \item[] \textbf{DOI}: \url{https://doi.org/10.5281/zenodo.14533132}
    \item[] \textbf{URL}: \url{https://github.com/sciknoworg/OntoAligner}
\end{itemize}

\section{Introduction}
Ontology alignment (OA) is a critical task for achieving semantic interoperability among diverse knowledge systems.  Ontologies serve a key building block for many applications, such as database integration, knowledge graphs (KGs), e-commerce, semantic web services, or social networks. However, evolving systems within the semantic web generally adopt different ontologies~\cite{euzenat2007ontology}. Hence, OA, the process of identifying correspondences between entities in different ontologies, spans various domains, including semantic web services, biomedical research, and KG integration. Despite significant advancements in OA research (cf. \cite{portisch2022background,mohammadi2020evaluating}), practitioners still face challenges related to scalability, modularity, robustness, and reproducibility when developing or deploying alignment systems. Many existing tools struggle with scalability when processing large ontologies, lack modularity to accommodate diverse use cases, and fall short in integrating contemporary AI methodologies. Additionally, limited support for customization, coupled with sparse documentation, creates barriers to adoption in both academic and industrial contexts. These limitations underscore the urgent need for a robust, flexible, and user-friendly OA framework that can cater to the evolving demands of modern applications.

Building on previous investigations, we observed a prevailing trend in OA is the use of large language models (LLMs), as highlighted by several studies \cite{olala,he2023exploring,qiang2023agent,silva2024complex,amini2024towards,zhang2024large,sousa2025complex,taboada2025ontology}. However, addressing the challenges within OA frameworks requires more than simply leveraging LLMs. Motivated by our prior research on the empirical evaluation of LLMs for OA~\cite{babaei2024llms4om} and the growing interest in LLM-driven approaches (as a starting point), we introduce OntoAligner, a comprehensive Python-based toolkit designed to advance the state of OA tools. OntoAligner adopts a modular architecture that seamlessly integrates traditional heuristic methods, retrieval-based models, and cutting-edge LLMs such as (but not limited to) Mistral~\cite{jiang2023mistral7b}, LLaMA~\cite{touvron2023llama2openfoundation}, Falcon~\cite{almazrouei2023falconseriesopenlanguage}, and Qwen~\cite{qwen}. The toolkit provides a flexible framework for OA, allowing users to configure and customize alignment algorithms, incorporate new datasets, and fine-tune pipelines with minimal effort. This modularity ensures that OntoAligner can adapt to a wide range of use cases, from lightweight heuristic alignments to advanced scenarios leveraging retrieval-augmented generation (RAG)~\cite{gao2024retrievalaugmentedgenerationlargelanguage} and in-context learning (ICL)~\cite{dong2024surveyincontextlearning}. Furthermore, OntoAligner emphasizes scalability, efficiently handling large and complex ontologies by optimizing memory usage and computational performance.

OntoAligner introduces several key features that distinguish it from existing tools. It combines traditional methods like fuzzy matching with advanced techniques such as RAG and ICL to ensure high alignment quality. Its intuitive design supports both step-by-step workflows and integrated pipelines, enabling users to tailor alignment processes to their needs. The toolkit also provides comprehensive evaluation metrics, such as precision, recall, and F1-score, alongside robust post-processing methods to ensure the consistency of alignment outputs. OntoAligner facilitates extensibility, allowing researchers to integrate custom alignment algorithms, post-processing steps, and evaluation modules. Moreover, the toolkit’s extensive documentation, tutorials, and examples enhance its accessibility for users with varying levels of expertise. The growing complexity and scale of data in domains like KG integration~\cite{abu2021domain}, semantic search~\cite{shalagin2024survey}, and biomedical research~\cite{gonzalez2024landscape} have amplified the demand for OA tools that are accurate, scalable, and adaptable. Existing solutions often fail to meet these requirements, as they are either computationally expensive, limited in scope, or not well-suited for integration with modern AI technologies. OntoAligner addresses these gaps, providing a versatile and high-performance toolkit designed to meet the needs of contemporary OA tasks. The primary contribution of this paper is presented as follows:
\begin{itemize}
    \item We introduce OntoAligner, a modular and extensible framework that integrates lightweight and AI-driven methods for OA, setting a new benchmark for versatility and adaptability. Through rigorous experimentation on benchmark datasets, we validate its performance, demonstrating superior precision, recall, and scalability compared to state-of-the-art tools. To promote innovation and collaboration within the OA community, we release OntoAligner as an open-source resource (can be installed via \url{https://pypi.org/project/OntoAligner/}), complete with comprehensive documentation (\url{https://ontoaligner.readthedocs.io/}) and community support.
\end{itemize}
With OntoAligner, we aim to bridge the gap between cutting-edge research and practical application, empowering researchers and practitioners to advance the field of OA. We envision OntoAligner as a central hub for open-source OA systems, fostering innovation and collaboration within the field.

\section{Related Work}\label{sec:rw}
OA involves aligning concepts, relationships, and instances across different ontologies to ensure semantic interoperability. Several OA Python frameworks have been developed to address various aspects of OA, each with its own strengths and limitations.

DeepOnto~\cite{he2024deeponto} (from KRR-Oxford lab) is a deep learning-based ontology matching framework that integrates various machine learning models, such as BERTMap~\cite{he2022bertmap} and Ontology Alignment Evaluation Initiative (OAEI)~\cite{OAEI23,oaei2024} resources for Bio-ML~\cite{he2022machine} and Bio-LLM~\cite{he2023exploring} tracks.  DeepOnto is not a pure Python-based toolkit. It leverages the OWL API -- a Java library -- to initialize custom ontologies, making it flexible and adaptable for different alignment tasks. However, to integrate Java dependency, DeepOnto uses JPype to bridge Python and the Java Virtual Machine (JVM). This setup can lead to installation challenges on systems not configured for Java support. While DeepOnto excels in accuracy, particularly in the Bio-ML track of the OAEI 2022-2024 dataset, it requires significant computational resources, particularly for training. Additionally, users must manually generate train-test splits for datasets outside predefined ones, and the framework’s documentation is sparse, offering only basic installation instructions. Finally, it is a general-purpose framework that is not only for OA but for ontology engineering tasks. Despite the mentioned drawbacks, DeepOnto remains an actively developed tool with high accuracy in its specialized domain and a growing community of users contributing to its ongoing improvement. Similarly, Matcha-DL (from Liseda-Lab)~\cite{cotovio2024matcha} is another deep learning-based OA framework that focuses on reducing the need for labeled datasets through semi-supervised learning. It builds on the established AgreementMakerLight system and auto-tunes hyperparameters to improve usability. However, like DeepOnto, Matcha-DL requires a good GPU for training and lacks pre-trained models. It is actively developed, with recent updates, but its minimal documentation and lack of real-world examples make it challenging for new users. Despite these challenges, Matcha-DL’s auto-tuning feature and semi-supervised learning approach offer promising solutions for users seeking to reduce dependency on labeled data.

OntoEMMA ontology matcher~\cite{OntoEmma} (from AllenAI lab) provides a machine learning-based approach to OA, with a minimal hybrid method incorporated. Unlike DeepOnto, OntoEmma supports both CPU and GPU training, making it more accessible for users with less powerful hardware. However, it lacks customization options for advanced users and suffers from minimal documentation and a lack of maintenance, with the last update occurring four years ago, despite these limitations. On the other hand, the Alignment framework~\cite{okgreece_alignment} takes a more logic-based approach to OA, emphasizing reasoning and consistency checks. It supports social features like a user voting system to improve link sets, and these alignments are made accessible via a SPARQL endpoint and API. However, it does not incorporate machine learning or hybrid techniques and struggles with scaling, particularly when working with incomplete ontologies. While lightweight and easy to set up with a Docker image, the Alignment framework has been abandoned for several years, with the last update occurring four years ago. The framework is useful for those focused on logical alignments and for non-pythonic users, but its lack of updates and limited domain flexibility make it less appealing for more complex or evolving alignment tasks. Moreover, IBM’s Ontology Alignment framework~\cite{ibm2024ontologyalignment} targets business use cases, such as data integration, and relies on machine learning for ontology matching. It supports large datasets, making it suitable for enterprise-level applications. However, it lacks domain-specific models and pre-trained models, and its integration with other IBM tools complicates its setup. Despite these challenges, the framework is relatively modern compared to other tools, with the last update occurring just one year ago. The documentation is straightforward for basic usage but lacks depth for advanced user-specific documentation, and there are few code comments or detailed API references.


Overall, DeepOnto~\cite{he2024deeponto} and Matcha-DL~\cite{cotovio2024matcha} stand out in deep-learning-based approaches; however, both require resources for fine-tuning. Additionally, DeepOnto's reliance on a Java backend may lead to compatibility issues across different operating systems. Moreover, OntoEMMA~\cite{OntoEmma}, as one of the tools, offers scalability but lacks customization and is no longer maintained.  Similarly, Alignment framework~\cite{okgreece_alignment}, while lightweight and logic-based, is limited in its scalability and has not seen updates in years. IBM's Ontology Alignment framework~\cite{ibm2024ontologyalignment}, though modern, is complex to set up and it is not being maintained. While these frameworks offer valuable features for OA, the common theme is trade-offs between flexibility, ease of use, scalability, and documentation quality. However, with OntoAligner, you may use LLMs that require fewer resources, such as those 1B models, and obtain a better alignment with advanced methods, such as RAG, or use existing lightweight models. It is being maintained regularly to cope with the newest needs of Python developers. The early design principle aimed for its extendability and modularity in supporting diverse approaches for OA, ranging from machine learning-based solutions to logic-based methods, each aiming at different user needs and computational capabilities. While various frameworks provide valuable features for ontology alignment (OA), they often involve trade-offs related to flexibility, ease of use, scalability, and documentation quality. In contrast, OntoAligner allows you easy usage of advanced methods while achieving better alignment.

\section{OntoAligner Framework}
Requirements are the basis for any successful software development, including libraries. For this reason, we elicited requirements for our OA framework based on an analysis of the presented related work (cf. Section~\ref{sec:rw}) and our empirical investigation of LLMs~\cite{babaei2024llms4om}. In this way, we can build on proven features by incorporating them into our framework and, at the same time, document current deficits through specified requirements that must be implemented in a targeted manner in the long term. Below, we first present the requirements followed by their actual implementation.

\begin{table}[htb!]
\centering
\caption{Functional (F) and Non-Functional (N) Requirements for OntoAligner Python Toolkit for Ontology Alignment. The \textit{Proven Feature} column represents the successfully built and tested functionalities. But, the \textit{Current Deficit} column shows areas that need future development.}
\label{tab:reqs}
\resizebox{\textwidth}{!}{%
\begin{tabular}{|c|p{11.5cm}|c|c|}
\hline
 & \textbf{Requirement} & \textbf{Proven Feature} & \textbf{Current Deficit} \\ \hline

F1 & The library shall be able to load ontologies from various local file formats (e.g., OWL, RDF). &  \checkmark &  \\ \hline
F2 & The library will support loading ontologies from remote URLs. &  & \checkmark \\ \hline
F3 & The library must provide a set of predefined alignment algorithms, i.e., lightweight-based, retrieval-based, LLM-based, RAG-based, and in-context learning-based algorithms. & \checkmark &  \\ \hline
F4 & The library should provide users with the ability to select and configure alignment algorithms based on their needs. & \checkmark  &  \\ \hline
F5 & The library must be able to generate mappings between classes of the source and target ontology. & \checkmark  &  \\ \hline
F6 & The library should provide the user with the ability to generate mappings between properties and individuals. &  & \checkmark \\ \hline
F7 & Model-specific post-processing is required to ensure the consistency of outputs. &  \checkmark &  \\ \hline
F8 & The library must provide the user with the ability to export mappings in a standard format, i.e., XML and JSON. & \checkmark &  \\ \hline 
F9 & The library must provide metrics to evaluate the quality of the generated mappings, i.e., precision, recall, and F-measure. & \checkmark &  \\ \hline
F10 & The library should provide the user with the ability to compare mappings generated by different algorithms or configurations. &  \checkmark & \\ \hline
F11 & The library should provide the developer with the ability to add new alignment algorithms or modify existing ones. &  \checkmark &  \\ \hline
F12 & The library should be well-documented for extending its functionalities. & \checkmark  &  \\ \hline\hline
N1 & The library should be able to align large ontologies, i.e., containing thousands of classes and properties, within a reasonable time frame. & \checkmark &  \\ \hline
N2 & The library should optimize memory usage to handle large ontologies efficiently. & \checkmark &  \\ \hline
N3 & The library should be intuitive and easy to use, with clear documentation and error messages. & \checkmark &  \\ \hline
N4 & The library should provide the user with the ability to learn how to use it by means of comprehensive user guides, examples, and tutorials. &  \checkmark &  \\ \hline
N5 & The library will be able to handle exceptions gracefully by providing meaningful error messages to the user. &  & \checkmark \\ \hline
N6 & The library should ensure the consistency of the generated mappings. & \checkmark &  \\ \hline
N7 & The library should be able to handle increasing amounts of data and users efficiently. & \checkmark &  \\ \hline
N8 & The library should be well-documented and modular to facilitate maintenance and future enhancements. & \checkmark &  \\ \hline
N9 & The library should follow best practices for software development, including version control and testing. & \checkmark &  \\ \hline

\end{tabular}%
}
\end{table}

\subsection{Requirements Specification}
The following elicited requirements represent key features, functionalities, and quality characteristics for an OA framework. These requirements can serve as guidance for us and other developers to ensure the targeted development of a high-quality OA framework. We used the established MASTER template by Chris Rupp and The SOPHISTs~\cite{Rupp.2021,TheSophists.2016} to document functional and non-functional requirements. The benefit of this template is an improved quality of requirements documentation by providing a structured approach that ensures clear, consistent, complete, and testable requirements of high quality. In particular, the template offers the advantage of defining the degree of legal obligation using the three keywords ``shall'', ``should'', and ``will''.  The ``shall'' indicates a mandatory requirement that must be implemented, “should” indicates an optional requirement whose implementation can significantly improve the software, and “will” indicates planned requirements for future integration that go beyond the current scope of the software.

The requirements are presented in \autoref{tab:reqs}, where functional requirements (denoted by "F") describe essential features and capabilities of the OntoAligner Python Toolkit for OA, while non-functional requirements (denoted by "N") outline performance, usability, and other quality-related aspects of the software. These requirements help ensure that the OntoAligner library can be effectively used, extended, and maintained while fulfilling the key goals of OA. By clearly defining these requirements, we provide a structured approach to both development and quality assurance. The \textit{Proven Feature} column in the \autoref{tab:reqs} reflects the functionalities that have already been implemented or verified, whereas the \textit{Current Deficit} column highlights areas where additional work is needed or where improvements are planned for future releases. Together, \textit{Proven Feature} and \textit{Current Deficit} columns allow for effective tracking of progress and identification of any gaps, ensuring that the toolkit evolves in line with user needs and expectations.

With respect to the standard requirements specification for software engineering, as detailed in \autoref{tab:reqs}, OntoAligner fulfills 10 out of 12 functional requirements. Additionally, it successfully meets 8 out of 9 non-functional requirements. This adherence to the requirements underscores the tool’s suitability for diverse applications in semantic alignment and data integration.

\subsection{Software Implementation}
\begin{figure}[t]
    \centering
    \includegraphics[width=\textwidth]{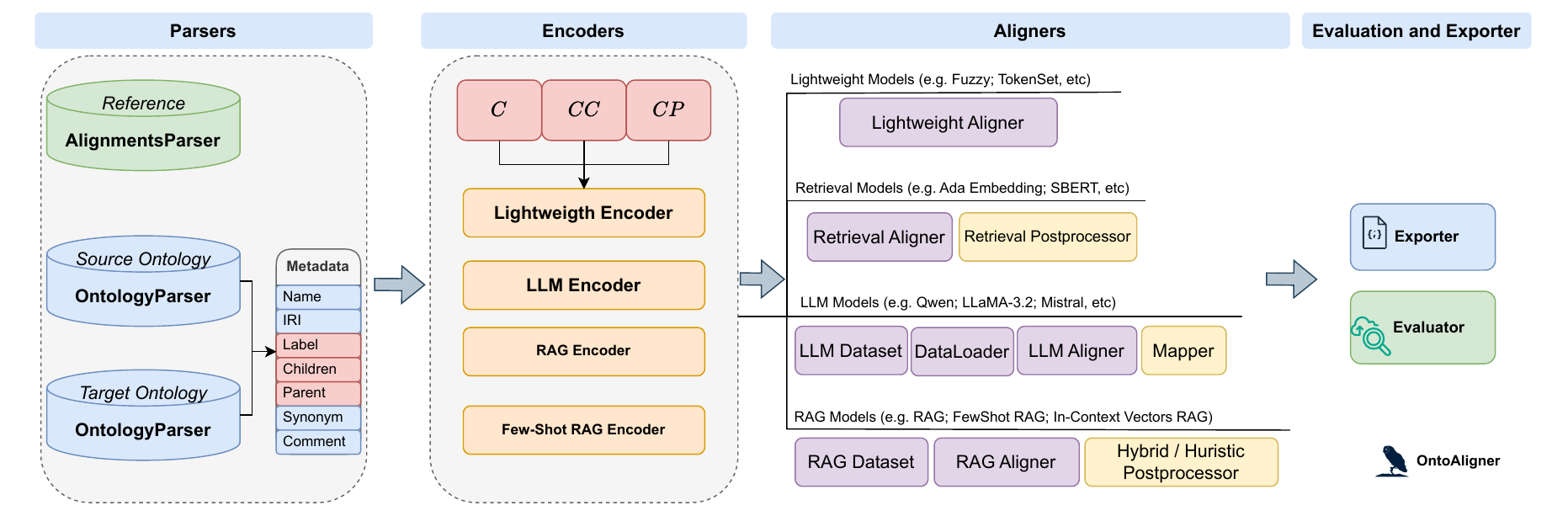}
    \caption{OntoAligner Architecture.}
    \label{fig:arch}
\end{figure}

The OntoAligner is implemented in a Python programming language to facilitate OA through a modular and extensible architecture. The system adheres to the functional and non-functional requirements outlined during the requirements engineering phase, ensuring robust, scalable, and user-friendly operation. The OntoAligner architecture is represented in \autoref{fig:arch} as a high-level abstraction of design consideration that has been taken into account during implementation. 
We detail the implementation of the core components of OntoAligner and their respective functionalities.

\noindent\textbf{1) Parser Module. }This module consists of two important components, \textit{Alignment} and \textit{Ontology} parsers. The ontology parsing module supports the ingestion of ontologies in standardized formats such as OWL and RDF, either from local files or remote URLs (F1, F2). The parsing process extracts essential metadata, including class and property names, IRIs, hierarchical structures, synonyms, and annotations, enabling compatibility with diverse ontology structures. However, advanced metadata (e.g., highly expressive OWL axioms) may require further processing or adaptations in future versions to support alignment algorithms with different metadata needs. This functionality is encapsulated in the \texttt{OntologyParser} class, which is the foundation for downstream alignment tasks. Moreover, the alignment parser encapsulated in the \texttt{AlignmentsParser} class loads ground truth alignments for OA tasks, which will later be used to evaluate alignment efficiency.

\noindent\textbf{2) Encoder Module. } After parsing the alignment ontologies, the framework employs multiple encoders to effectively represent ontological concepts in suitable data structures (e.g., natural language text or paired concepts for later similarity estimation) for subsequent processing. Drawing insights from the previous study on ontology representations in~\cite{babaei2024llms4om}, this module utilizes Concept $C$ (basic concept representation), Concept-Children $CC$ (lower-level taxonomy), and Concept-Parent $CP$ (upper-level taxonomy) as foundational metadata. These representations serve as the basis for the following encoders, which are designed as modular components and include:
\begin{itemize}
    \item  \underline{\textit{Lightweight Encoder.}} This utilizes forming inputs as a raw text, which is a simple lightweight approach for heuristic techniques such as Fuzzy Matching and Retrieval Matching for computationally inexpensive alignments. For $C$, $CC$, and $CP$, it just combines concepts with respective pairs in a natural language text. 
    \item \underline{\textit{LLM Encoder.}} It uses the desired representation to clean the text and obtain parents and children in desirable formatting (each concept and its combined parents stored in a dictionary) for integrating into prompt placeholders for inferences with LLMs.
    \item \underline{\textit{RAG Encoder.}} Aims to provide structured inputs for both retriever and LLMs in leveraging  RAG~\cite{rag} with Few-Shot and ICL-- specifically we incorporated the in-context vectors (ICV) approach introduced by Liu et al.~\cite{liu2024context} as an aligner which requires encoder -- for advanced alignment scenarios. 
\end{itemize}
This modular design (F3) allows users to choose and configure the most suitable encoder for their specific use case (F4). 


\noindent\textbf{3) Aligner Module. }The core alignment process is performed by the \textit{Aligners} module for the alignment algorithm, which includes several predefined categories:
\begin{itemize}
    \item \underline{\textit{Lightweight Aligners.}} Fast and heuristic-based algorithms for mappings between entities. Lightweight aligners include fuzzy matchings~\cite{fuzzywuzzy} such as simple token sets and weighted fuzzy matching models that take matching thresholds to perform alignments. 
    \item \underline{\textit{Retrieval Aligners}}. A machine learning model is designed to retrieve matches based on semantic similarity or keyword matching. They mostly operate on semantic similarity techniques using vector representations. Retrieval aligners within OntoAligner consist of sentence-transformer~\cite{sbert}-based, TFIDF, and SVM~\cite{karpathy_knn_svm} models. The sentence-transformer module is compatible with a wide range of transformer models from Hugging Face, including both \href{https://huggingface.co/sentence-transformers}{official models} and \href{https://huggingface.co/models?library=sentence-transformers}{community-contributed} models.
    \item \underline{\textit{LLM Aligners}}. An LLM-based approach that leverages LLMs from the Hugging Face repository for alignment. These aligners operate using predefined prompts to query LLMs (the prompts have been discussed within \cite{babaei2024llms4om}), determining whether a match exists between pairs of concepts. This set of aligners operates with quadratic time complexity but remains practical for smaller ontology alignment tasks involving fewer than $\approx$200 concepts in the source and target ontologies.
    \item \underline{\textit{RAG Aligners}}. A hybrid alignment method that integrates retrieval and generative capabilities of LLMs to produce high-quality alignments. These aligners retrieve relevant knowledge before generating alignment decisions, enhancing accuracy and contextual understanding. On the LLM side, instead of relying on full-text generation, OntoAligner utilizes logit-based probability calculations to determine alignments. This reduces the token sampling overhead associated with text generation, which allows them to operate on the single forward pass. This reduces GPU usage.
\end{itemize}
The framework allows users to configure these algorithms, ensuring flexibility and adaptability to various OA scenarios (F4). Moreover, the \textit{Post-Processing} steps are implemented to refine the initial mappings generated by the aligners. These include LLM-based mapping (mapping generated natural language to the pre-defined classes) and rule-based and heuristic filtering/processing adjustments to ensure the consistency and quality of alignments (F7, N6). The modular nature of the post-processing pipeline enables easy integration of additional refinement techniques. The post-processing module is being implemented as different components to support users' desirable post-processing algorithms for different models. The \texttt{Mapper} is a post-processing technique that is useful for assigning appropriate labels to generated LLM texts, especially in \textit{LLM Aligner}.

\noindent\textbf{4) Evaluation and Exporter Modules. }To assess the quality of generated alignments, the framework includes an \textit{Evaluator} module that computes standard metrics such as precision, recall, and F-measure (F9). Users can also compare mappings produced by different algorithms or configurations (F10). The final mappings can be exported in widely used formats, including XML and JSON, via the \textit{Exporter} module (F8). 

The OntoAligner framework is designed for extensibility, allowing developers to add custom alignment algorithms or modify existing ones through a well-documented API (F11, F12). The code base adheres to modular object-oriented design principles, promoting ease of maintenance and scalability (N8) while allowing base foundational modules for extensions based on newer alignment frameworks. To handle large ontologies efficiently, OntoAligner optimizes memory usage and computational performance, enabling the alignment of ontologies containing thousands of classes and properties within a reasonable time frame (N1, N2). Additionally, the framework includes good documentation, user guides, and tutorials to ensure accessibility for users of varying expertise levels (N3, N4). By implementing these components, OntoAligner maintains flexibility and ease of use, making it a valuable tool for researchers and practitioners.

\subsection{Documentation}
A comprehensive and detailed software documentation of OntoAligner is created via the read.the.docs. It can be accessed at \url{https://ontoaligner.readthedocs.io/index.html}. The documentation is organized into the following main parts: \textit{Getting Started}, \textit{How to Use?}, \textit{Aligners}, and \textit{Package References}. 

\noindent\textbf{Getting Started.} The \textit{Getting Started} section provides all the necessary information to help users begin using OntoAligner effectively. It includes detailed \textit{installation} instructions and a \textit{quickstart} guide that demonstrates how to set up and run OntoAligner for the first time, ensuring a smooth onboarding process. Moreover, the \textit{How to Use?} section offers a comprehensive overview of the OntoAligner’s features and functionalities. It includes guidance on navigating the models, where users can explore and choose the most suitable models for their ontology alignment needs or add their own LLM into action. 

\noindent\textbf{Aligners.} The \textit{Aligners} section is a kind of tutorial that delves into the various aligners available in OntoAligner. This includes lightweight aligners for basic tasks, retrieval-based aligners for enhanced matching, and advanced aligners like those using LLMs, RAG, and ICL. The step-by-step code explanation with outputs walks the users through the models to ensure ease of use. 

\noindent\textbf{Package Reference.} Finally, the \textit{Package Reference} section serves as a technical reference for OntoAligner’s modules and functions. It covers all key components, including the pipeline workflow, foundational base module, ontology representation tools, matching algorithms, encoders, post-processing techniques, and utility functions. This section provides in-depth insights into the internal workings of OntoAligner, empowering developers to customize its functionalities. 

\section{Evaluation}
In this section, we demonstrate a detailed assessment of the OntoAligner's capabilities and its effectiveness through designed tests and experimentation on multiple OA datasets. The statistics of case-study datasets are presented in \autoref{tab_dataset_stats} with respect to their representations and alignments.

\begin{table}[t]
    \centering
    \small
    \caption{OAEI tracks and records statistics for sources, targets, and alignments. "MI" refers to the "MaterialInformation" ontology, while "S" and "T" denote the source and target ontologies, respectively.}\label{tab_dataset_stats}
   \resizebox{\textwidth}{!}{
    \begin{tabular}{|l|l|r|r|r|r|r|r|r|}
        \hline
        \multirow{2}{*}{\textbf{Track}} & \multirow{2}{*}{\textbf{Task}} & \multicolumn{2}{|c|}{\textbf{Concepts}} &
        \multicolumn{2}{|c|}{\textbf{Children}} & \multicolumn{2}{|c|}{\textbf{Parents}} &
        \multirow{2}{*}{\textbf{Alig}}\\
        \cline{3-8}
         &  & \multirow{1}{*}{S} & \multirow{1}{*}{T} & \multirow{1}{*}{S} & \multirow{1}{*}{T} & \multirow{1}{*}{S} & \multirow{1}{*}{T} & \\
        \hline
         \hline
        \multirow{1}{*}{\textsc{Anatomy}~\cite{dragisic2017experiences}}& Mouse-Human & 2,737&3,298 &482&673 & 1,687&3,297 & 1,516\\
        \hline
        \multirow{2}{*}{\textsc{Biodiv}~\cite{karam2020matching}}&  FISH-ZOOPLANKTON & 145&56 &145&56 & 34&7 & 15\\
        & ALGAE-ZOOBENTHOS & 108&128 &108&123 & 24&27 & 18\\
        \hline\multirow{1}{*}{\textsc{Phenotype}~\cite{harrow2017matching}}& HP-MP & 12,786&11,928 &4,387&4,439 & 12,646&11,498 & 696\\
        \hline\multirow{2}{*}{\textsc{CommonKG}~\cite{fallatah2020gold}}& Nell-DBpedia & 134&137 &0&0 & 0&0 & 129\\
        & Yago-Wikidata & 304&304 &0&0 & 0&0& 304\\
        \hline\multirow{2}{*}{\textsc{MSE}~\cite{nas2023mse}}& MI-EMMO & 545&903 &64&232 & 536&704 &63\\
        & MI-MatOnto & 545&825 &64&114 & 536&793 & 302\\
        \hline
    \end{tabular}
    }
\end{table}

\begin{figure}[htb!]
\centering
\includegraphics[width=\textwidth]{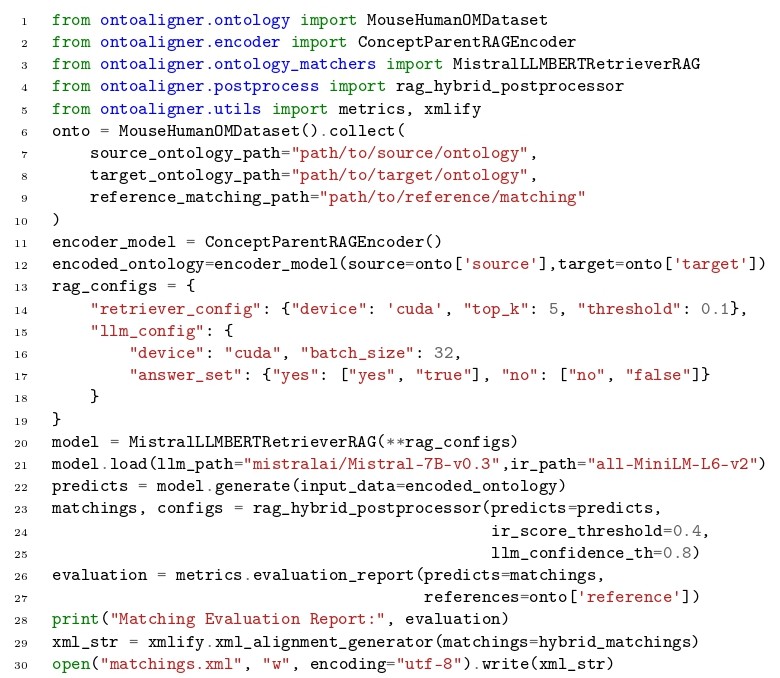}
\caption{OntoAligner step-by-step code for RAG-based OA.}
\label{lst:ontoaligner-rag-code}
\end{figure}

\begin{figure}[htb!]
\centering
\includegraphics[width=\textwidth]{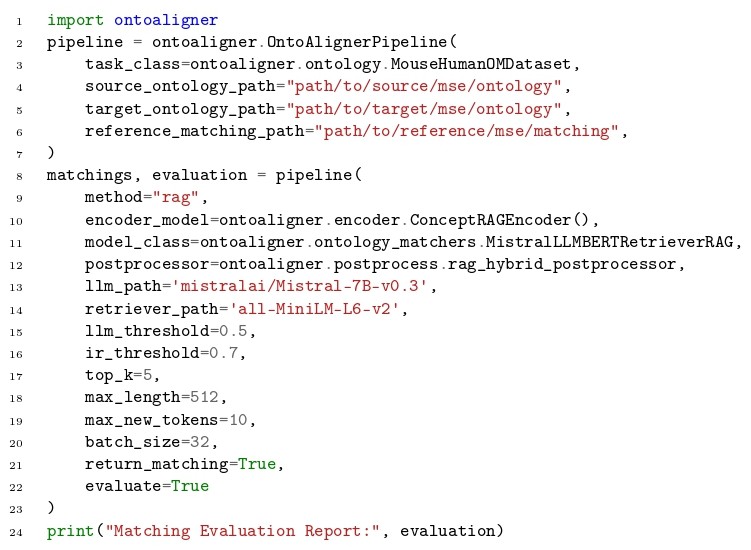}
\caption{Python code for Pipeline using the OntoAligner library.}
\label{lst:ontoaligner-pipeline-code}
\end{figure}

\subsection{User Flexibility and Customization in OntoAligner}
\noindent\textbf{Ease of Use.} OntoAligner demonstrates a modular and extensible design, enabling users to customize and integrate components seamlessly. 
\autoref{lst:ontoaligner-rag-code} presents the step-by-step implementation of OntoAligner for RAG-based OA in just 30 lines of code. Lines 6–10 load the ontology dataset, while lines 11–12 encode it using a concept-parent-based RAG encoder. Lines 13–19 define the retriever and LLM configurations, followed by lines 20–22, which load the RAG model and generate matchings, and post-processing with threshold filtering is applied in lines 23–25. Finally, lines 26–28 evaluate the results against reference alignments, and lines 29 convert the matchings into an XML format. This highlights the modularity of OntoAligner, making it easy to configure and reuse components. The granular control allows advanced users to tailor the alignment pipeline to specific needs. In contrast, \autoref{lst:ontoaligner-pipeline-code} showcases how these modular components are elegantly orchestrated into a cohesive pipeline, promoting code reusability and maintainability. By offering both step-by-step and pipeline-based workflows, OntoAligner strikes a balance between flexibility and simplicity, making it a robust and well-architected solution for OA.



\noindent\textbf{Parameter Customization with OntoAligner. }\autoref{tab:performance-comparison} presents the results of the OA experimentation conducted on the MaterialInformation-MatOnto task from the MSE track~\cite{nasr2020evaluation} (the second task of MSE) of the OAEI~\cite{OAEI23}. In this MSE task, the source ontology consists of 545 concepts, while the target ontology contains 825 concepts. The ground truth includes a total of 302 alignments. The best-performing method reported for this task in the OAEI 2023 competition was Matcha~\cite{faria2023results}, which achieved an F1-score of 33.9\% with a response time of 15 seconds. Our experimentation with the OntoAligner library demonstrates that by leveraging its modular design, parameter flexibility, and advanced RAG techniques, we can achieve substantial improvements over the state-of-the-art. The experimentation highlights the power of FewShot RAG and ICV RAG methods combined with the sentence-transformer-based retrieval model with Mistral-7B or Falcon-7B LLMs. By systematically customizing the key parameters such as retriever thresholds ($T_r$), LLM thresholds ($T_l$), top-k retrieval values, and batch sizes ($B$), we achieved an F1-score of 44.1\%, surpassing the Matcha baseline by over 10 percentage points. Notably, the FewShot RAG approach using SBERT(C) with Mistral-7B(CC) achieved this performance with precision and recall values of 64.7\% and 33.4\%, respectively, showcasing a significant tradeoff optimization between precision and recall.

\begin{table}[htb!]
    \centering
    \caption{Result for OA experimentation using various tools from OntoAligner for MaterialInformation-MatOnto task (the second task) from the MSE‌ track~\cite{nasr2020evaluation} of the Ontology Alignment Evaluation Initiative. The best performer for the task is Matcha~\cite{faria2023results}, with an F1-Score of 33.9\% and a response time of 15 seconds. In the table, results are in percentages, with "P" referring to precision and "R" referring to Recall metrics. Time refers to the response time for the method, and it is in seconds. "MI" in "MI-MatOnto" and "MI-EMMO" datasets refers to the "MaterialInformation" ontology. \textit{ICV RAG} refers to in-context-vectors (ICV) -based RAG models, where ICV is introduced in \cite{liu2024context}. The bold score in the F1 column refers to the best performance obtained using OntoAligner.}
    \label{tab:performance-comparison}
    \resizebox{\textwidth}{!}{
    \begin{tabular}{|l|l|r|r|r|r|l|}
        \hline
\textbf{Method}  & \textbf{Model} (\textbf{Encoder}) & \textbf{P} &\textbf{ R} & \textbf{F1} & \textbf{Time} & \textbf{Parameters} \\\hline \hline
\textit{\textbf{Lightweight}} &  SimpleFuzzy (C)  & 14.3 & 25.8 & 18.4  & 0.13 & $T=0.1$  \\ \hline
\textit{\textbf{Retrieval}} &   SBERT (MiniLM) (C) & 3.3 & 60.9 & 6.3 & 8.10 & $top_k=10$,$T=0.2$ \\ \hline
\textit{\textbf{Retrieval}} &  SBERT (MiniLM) (CC) & 1.8 & 67.5 & 3.6 & 3.38 & $top_k=20$, $T=0.2$ \\ \hline
\textit{\textbf{Retrieval}} &  TFIDF (CP) &  1.1 & 27.4 & 2.2 & 1.91 & $top_k=20$, $T=0.2$  \\ \hline
\textit{\textbf{LLM}} & Qwen2-0.5B (C) & 0.1 & 5.6  & 0.2  & 1065.5 & TFIDFLabelMapper, $B=2048$ \\ \hline
\textit{\textbf{LLM}} & LLaMA-3.2-1B (C) &  0.0 & 23.5 & 0.13  & 1441.3 & TFIDFLabelMapper, $B=1024$ \\ \hline
\textit{\textbf{RAG}} &SBERT(C) + Mistral-7B(C)  & 42.7 & 32.4  & 36.9  & 115.2  & $T_{l}=0.6$, $T_{r}=0.4$,$top_k=5$, $B=64$  \\ \hline
\textit{\textbf{FewShot RAG}} & SBERT(C) + Mistral-7B(C)  & 71.2 & 25.4 & 37.5  & 172  & $T_{l}=0.6$, $T_{r}=0.4$, $n_{s}=2$,$top_k=5$, $B=64$  \\ \hline
\textit{\textbf{FewShot RAG}} & SBERT(C) + Mistral-7B(C)  & 63.8 &  32.1  & 42.7 & 175 & $T_{l}=0.4$, $T_{r}=0.4$, $n_{s}=2$,$top_k=5$, $B=64$  \\ \hline
\textit{\textbf{FewShot RAG}} & SBERT(C) + Mistral-7B(CC)  &  64.7  & 33.4   &  \textbf{44.1} & 352.4 & $T_{l}=0.4$, $T_{r}=0.4$, $n_{s}=2$,$top_k=5$, $B=32$  \\ \hline
\textit{\textbf{FewShot RAG}} & SBERT(C) + Mistral-7B(CP)  & 84.7   & 25.8 & 39.5 & 201.9 & $T_{l}=0.4$, $T_{r}=0.4$, $n_{s}=2$,$top_k=5$, $B=64$  \\ \hline
\textit{\textbf{ICV RAG}} & SBERT(C) + Falcon-7B(C)  & 38.9  &  32.1  &  35.2  &  82.2 & $T_{l}=0.4$, $T_{r}=0.4$, $B=64$   \\ \hline
\textit{\textbf{ICV RAG}} & SBERT(C) + Falcon-7B(CC)  & 38.5 & 31.7  & 34.8  &  175.1 & $T_{l}=0.4$, $T_{r}=0.4$, $B=64$   \\ \hline
\textit{\textbf{ICV RAG}} & SBERT(C) + Falcon-7B(CP)  & 38.9 & 32.1  & 35.2  &   93.4 & $T_{l}=0.4$, $T_{r}=0.4$, $B=64$   \\ \hline
\end{tabular}
}
\end{table}

\subsection{Experimental Results}

\begin{table}[htb!]
\centering
\caption{Performance comparison across datasets and latest LLMs capabilities in six alignment tasks. In the table, results are in percentages with "Inter" referring to the number of intersections of predicted results with references,  "P" referring to precision metric,  "R" referring to Recall metric, "Pred" referring to the number of predicted matching by model, and "Ref" is referring to the number of gold matchings. Time refers to the response time for the method in seconds. The OAEI column refers to the best model results reported in OAEI.}
\label{tab:performances}
\resizebox{\textwidth}{!}{
\begin{tabular}{|l|l|l|l|l|l|l|l|l|l|}
\hline
\textbf{Dataset}          & \textbf{LLM (Retriever Model)}            & \textbf{Inter} & \textbf{P} & \textbf{R} & \textbf{F1} & \textbf{Pred} & \textbf{Ref} & \textbf{Time} & \textbf{OAEI (F1)}\\ \hline \hline
\textit{MI-MatOnto} & Mistral-7B-v0.3 (BERT)     & 102                 & 65.3          & 33.7        & \textbf{44.5}        & 156                      & 302                    & 342.7      &  33.9 (Matcha~\cite{faria2023results})    \\ \hline
\textit{MI-EMMO} & Qwen2-0.5B (BERT)     & 61                 & 88.4          & 96.8        & \textbf{92.4}        & 69                      & 63                    & 135.0      &  91.8 (Matcha~\cite{faria2023results})    \\ \hline
\textit{FISH-ZOOPLANKTON}  & LLaMA-3.2-1B (BERT)        & 13                  & 92.8            & 86.6         & \textbf{89.6}         & 14                       & 15                     & 9.9   & 64.0 (LogMapLt~\cite{logmap})         \\ \hline
\textit{Mouse-Human}    & LLaMA-3.2-3B  (BERT)  & 1291     & 87.7  & 85.1   & 86.4  & 1472   & 1516     & 116.5    & \textbf{94.1} (Matcha~\cite{faria2023results})  \\ \hline
\textit{ALGAE-ZOOBENTHOS}    &  Qwen2-0.5B (TFIDF)   &   12    &  75.0  & 66.6 &  \textbf{70.5}  & 16    &    18   & 2.80 & 44.4 (LogMapLt~\cite{logmap})  \\ \hline
\textit{Nell-DBpedia}    &  Qwen2-0.5B (T5)      &   126     &  97.6   & 97.6 &   \textbf{97.6}  &   129   &     129   &  5.9 & 96.0 (OLaLa~\cite{olala})  \\ \hline
\textit{YAGO-Wikidata}    &  Ministral-3B-Instruct (T5)    &    283     &   99.2   & 93.0  &  \textbf{96.0}   &   285    &   304      &  33.7  & 94.0 (Matcha~\cite{faria2023results})  \\ \hline
\textit{HP-MP}    &  Qwen2-0.5 (BERT)    &    667     &   74.1   & 95.8  &  \textbf{83.5}   &   900    &   696      &  2229.6  & 81.8 (LogMap~\cite{logmap})  \\ \hline
\end{tabular}
}
\end{table}

The results of experimentation\footnote{The code for experimentation is available here \url{https://github.com/sciknoworg/OntoAligner/blob/main/examples/OntoAlignerPipeline-Exp.ipynb}.} on eight OA datasets and various LLMs with different retrievers are presented in \autoref{tab:performances}. As we can see, according to the results, OntoAligner demonstrates exceptional performance in OA, consistently achieving high precision, recall, and F1 scores across diverse datasets, often surpassing state-of-the-art models reported in the OAEI benchmark. Notably, it achieves a close to perfect F1-score of 97.6\% on the Nell-DBpedia dataset and a strong 96.0\% on YAGO-Wikidata, outperforming top models like OLaLa~\cite{olala} and Matcha~\cite{faria2023results}. Its ability to identify meaningful matches is evident from the high number of intersections, as seen in datasets like Mouse-Human with 1291 intersections and an F1-score of 86.4\%. Additionally, the tool exhibits remarkable computational efficiency, delivering results in seconds for smaller datasets like ALGAE-ZOOBENTHOS while maintaining robust performance for larger datasets. These results establish OntoAligner as a powerful and reliable tool, pushing the boundaries of OA beyond current leading models. What makes OntoAligner a particularly promising tool for OA is its versatility across different LLMs, retrieval models, and datasets, showcasing adaptability to varying alignment tasks. The integration of the newest advanced LLMs, such as Mistral, LLaMA-3.2, and Qwen2.5, ensures high-quality predictions, as seen with models like \href{https://huggingface.co/ministral/Ministral-3b-instruct}{Ministral-3B-Instruct} achieving superior performance in precision and recall for the YAGO-Wikidata dataset. 

The lower performance of OntoAligner compared to Matcha in the Mouse-Human dataset is likely due to the differences in models. Matcha uses a fine-tuned encoder-decoder-LSTM for sequence-to-sequence alignment, while OntoAligner uses an RAG model that combines retrievals with LLMs. Since the RAG model is not domain-specifically fine-tuned, it may struggle with complex ontology concepts. Fine-tuning the RAG model or integrating more specialized mechanisms could improve future performance. Additionally, OntoAligner’s LLaMA-3.2-3B model can be run via Google Colab GPUs, whereas Matcha requires a training step that might require datasets and also resources for fine-tuning.

OntoAligner is designed to become a hub for diverse OA, allowing seamless integration without technical complexity. The results in \autoref{tab:performances} show how the integration of various LLMs can be tested with OntoAligner. The comparison of OntoAligner with top-performing tools, including Matcha (the backbone of Matcha-DL) and OLaLa (an RAG-based Java model using LLaMA-2-70B) contrasts in terms of size with our Qwen2-0.5B very small model, which performs well in Nell-DBpedia, HP-MP, and MaterialInformation-EMMO datasets. Our evaluation includes large datasets such as Mouse-Human (2,700 source, 3,300 target ontology concepts), MaterialInformation-EMMO (545 sources, 903 target ontology concepts), and HP-MP (12.7k source, 11.9k target ontology concepts). This supports the Scalability of OntoAligner to larger datasets.

\section{Future Work}
The project's maintenance plan, available at \url{https://github.com/sciknoworg/OntoAligner/blob/main/MAINTANANCE.md}, provides guidelines for ongoing development and support. Furthermore, the future work of OntoAligner will focus on three key aspects:

\noindent\textbf{Extending OntoAligner.} The OntoAligner library has been designed with extensibility and community collaboration at its core. In the immediate future, we aim to extend the tool by integrating additional state-of-the-art OA techniques such as LogMap~\cite{logmap}, Matcha~\cite{faria2023results}, and many more, including methods from lightweight OA to Agent OA approaches. Furthermore, as OntoAligner continues to evolve, it will support a wider range of alignment encoders tailored to different models, ensuring compatibility with diverse input formats and ontology structures while adhering to best practices in ontology processing.

\noindent\textbf{Community Collaboration.} We are also committed to fostering a vibrant community around OntoAligner. To this end, we actively invite collaboration from researchers, developers, and practitioners. Contributions can take many forms, such as suggesting improvements, reporting issues, or directly contributing code and documentation. By engaging with the community, we aim to ensure that OntoAligner remains relevant, user-friendly, and adaptable to evolving needs by extending the requirement specification.

\noindent\textbf{Promoting OntoAligner.} To promote wider adoption, we plan to showcase OntoAligner in various academic and professional workshops, including its integration into the Ontology Alignment Evaluation Initiative (OAEI). This will highlight the tool's capabilities while facilitating its uptake within the broader OA community. Additionally, these efforts will emphasize the importance of reusability and best practices in the development of alignment tools, further aligning OntoAligner with community standards and expectations.

\section{Conclusion}
In this paper, we presented OntoAligner, a novel Python library for ontology alignment. The proposed tool addresses ontology alignment tasks by leveraging lightweight, retriever, and advanced LLM models for a customizable alignment approach. Through comprehensive evaluation and use-case demonstrations, we showed that OntoAligner can achieve state-of-the-art performance in various OA tasks while maintaining flexibility and ease of use. The contributions of this work are not only in its technical advancements but also in its potential to enhance interoperability, promote knowledge reuse, and support practitioners in their research. By prioritizing ease of integration and community-driven development, we have ensured that OntoAligner is accessible and adaptable to various needs. Looking ahead, we aim to extend the functionality of OntoAligner and engage with the wider community for collaborative development. We believe that OntoAligner will serve as a valuable resource for researchers, developers, and practitioners in the field, fostering innovation and advancing the state of the art in ontology alignment and semantic web technologies.

\begin{credits}
\subsubsection{\ackname} We sincerely thank \textit{Mahsa Sanaei} and \textit{Amirreza Alasti} for their contributions to the codebase and documentation of the OntoAligner. This work is jointly supported by the \href{https://scinext-project.github.io/}{SCINEXT project} (BMBF, German Federal Ministry of Education and Research, Grant ID: 01lS22070), the KISSKI AI Service Center (BMBF, Grant ID: 01IS22093C), \href{https://www.nfdi4datascience.de/}{NFDI4DataScience} and \href{https://nfdi4ing.de/}{NFDI4ING} initiatives (DFG, German Research Foundation, Grant ID: 460234259 and 442146713).



\subsubsection{\discintname}
The authors have no competing interests to declare that are relevant to the content of this article.
\end{credits}

\bibliographystyle{splncs04}
\bibliography{main}

\begin{thebibliography}{10}
\providecommand{\url}[1]{\texttt{#1}}
\providecommand{\urlprefix}{URL }
\providecommand{\doi}[1]{https://doi.org/#1}

\bibitem{abu2021domain}
Abu-Salih, B.: Domain-specific knowledge graphs: A survey. Journal of Network and Computer Applications  \textbf{185},  103076 (2021)

\bibitem{OntoEmma}
{Allen Institute for Artificial Intelligence}: Ontoemma ontology matcher. \url{https://github.com/allenai/ontoemma} (2020)

\bibitem{almazrouei2023falconseriesopenlanguage}
Almazrouei, E., Alobeidli, H., Alshamsi, A., Cappelli, A., Cojocaru, R., Debbah, M., Étienne Goffinet, Hesslow, D., Launay, J., Malartic, Q., Mazzotta, D., Noune, B., Pannier, B., Penedo, G.: The falcon series of open language models (2023), \url{https://arxiv.org/abs/2311.16867}

\bibitem{amini2024towards}
Amini, R., Norouzi, S.S., Hitzler, P., Amini, R.: Towards complex ontology alignment using large language models. In: International Knowledge Graph and Semantic Web Conference. pp. 17--31. Springer (2024)

\bibitem{babaei2024llms4om}
Babaei~Giglou, H., D’Souza, J., Engel, F., Auer, S.: Llms4om: Matching ontologies with large language models. In: European Semantic Web Conference. pp. 25--35. Springer (2024)

\bibitem{qwen}
Bai, J., Bai, S., Chu, Y., Cui, Z., Dang, K., Deng, X., Fan, Y., Ge, W., Han, Y., Huang, F., Hui, B., Ji, L., Li, M., Lin, J., Lin, R., Liu, D., Liu, G., Lu, C., Lu, K., Ma, J., Men, R., Ren, X., Ren, X., Tan, C., Tan, S., Tu, J., Wang, P., Wang, S., Wang, W., Wu, S., Xu, B., Xu, J., Yang, A., Yang, H., Yang, J., Yang, S., Yao, Y., Yu, B., Yuan, H., Yuan, Z., Zhang, J., Zhang, X., Zhang, Y., Zhang, Z., Zhou, C., Zhou, J., Zhou, X., Zhu, T.: Qwen technical report. arXiv preprint arXiv:2309.16609  (2023)

\bibitem{cotovio2024matcha}
Cotovio, P.G., Ferraz, L., Faria, D., Balbi, L., Silva, M.C., Pesquita, C.: Matcha-dl a tool for supervised ontology alignment. preprint  (2024)

\bibitem{dong2024surveyincontextlearning}
Dong, Q., Li, L., Dai, D., Zheng, C., Ma, J., Li, R., Xia, H., Xu, J., Wu, Z., Liu, T., Chang, B., Sun, X., Li, L., Sui, Z.: A survey on in-context learning (2024), \url{https://arxiv.org/abs/2301.00234}

\bibitem{dragisic2017experiences}
Dragisic, Z., Ivanova, V., Li, H., Lambrix, P.: Experiences from the anatomy track in the ontology alignment evaluation initiative. Journal of biomedical semantics  \textbf{8},  1--28 (2017)

\bibitem{euzenat2007ontology}
Euzenat, J., Shvaiko, P., et~al.: Ontology matching, vol.~18. Springer (2007)

\bibitem{fallatah2020gold}
Fallatah, O., Zhang, Z., Hopfgartner, F.: A gold standard dataset for large knowledge graphs matching. In: Ontology Matching 2020: Proceedings of the 15th International Workshop on Ontology Matching co-located with the 19th International Semantic Web Conference (ISWC 2020). vol.~2788, pp. 24--35. CEUR Workshop Proceedings (2020)

\bibitem{faria2023results}
Faria, D., Silva, M.C., Cotovio, P., Ferraz, L., Balbi, L., Pesquita, C.: Results for matcha and matcha-dl in oaei 2023. In: OM@ ISWC. pp. 164--169 (2023)

\bibitem{gao2024retrievalaugmentedgenerationlargelanguage}
Gao, Y., Xiong, Y., Gao, X., Jia, K., Pan, J., Bi, Y., Dai, Y., Sun, J., Wang, M., Wang, H.: Retrieval-augmented generation for large language models: A survey (2024), \url{https://arxiv.org/abs/2312.10997}

\bibitem{gonzalez2024landscape}
Gonz{\'a}lez-M{\'a}rquez, R., Schmidt, L., Schmidt, B.M., Berens, P., Kobak, D.: The landscape of biomedical research. Patterns  (2024)

\bibitem{harrow2017matching}
Harrow, I., Jim{\'e}nez-Ruiz, E., Splendiani, A., Romacker, M., Woollard, P., Markel, S., Alam-Faruque, Y., Koch, M., Malone, J., Waaler, A.: Matching disease and phenotype ontologies in the ontology alignment evaluation initiative. Journal of biomedical semantics  \textbf{8},  1--13 (2017)

\bibitem{he2022bertmap}
He, Y., Chen, J., Antonyrajah, D., Horrocks, I.: Bertmap: a bert-based ontology alignment system. In: Proceedings of the AAAI Conference on Artificial Intelligence. vol.~36, pp. 5684--5691 (2022)

\bibitem{he2023exploring}
He, Y., Chen, J., Dong, H., Horrocks, I.: Exploring large language models for ontology alignment. arXiv preprint arXiv:2309.07172  (2023)

\bibitem{he2024deeponto}
He, Y., Chen, J., Dong, H., Horrocks, I., Allocca, C., Kim, T., Sapkota, B.: Deeponto: A python package for ontology engineering with deep learning. Semantic Web  \textbf{15}(5),  1991--2004 (2024)

\bibitem{he2022machine}
He, Y., Chen, J., Dong, H., Jim{\'e}nez-Ruiz, E., Hadian, A., Horrocks, I.: Machine learning-friendly biomedical datasets for equivalence and subsumption ontology matching. In: International Semantic Web Conference. pp. 575--591. Springer (2022)

\bibitem{olala}
Hertling, S., Paulheim, H.: Olala: Ontology matching with large language models. In: Proceedings of the 12th Knowledge Capture Conference 2023. p. 131–139. K-CAP '23, Association for Computing Machinery, New York, NY, USA (2023). \doi{10.1145/3587259.3627571}, \url{https://doi.org/10.1145/3587259.3627571}

\bibitem{jiang2023mistral7b}
Jiang, A.Q., Sablayrolles, A., Mensch, A., Bamford, C., Chaplot, D.S., de~las Casas, D., Bressand, F., Lengyel, G., Lample, G., Saulnier, L., Lavaud, L.R., Lachaux, M.A., Stock, P., Scao, T.L., Lavril, T., Wang, T., Lacroix, T., Sayed, W.E.: Mistral 7b (2023), \url{https://arxiv.org/abs/2310.06825}

\bibitem{logmap}
Jim{\'e}nez-Ruiz, E., Cuenca~Grau, B.: Logmap: Logic-based and scalable ontology matching. In: Aroyo, L., Welty, C., Alani, H., Taylor, J., Bernstein, A., Kagal, L., Noy, N., Blomqvist, E. (eds.) The Semantic Web -- ISWC 2011. pp. 273--288. Springer Berlin Heidelberg, Berlin, Heidelberg (2011)

\bibitem{karam2020matching}
Karam, N., Khiat, A., Algergawy, A., Sattler, M., Weiland, C., Schmidt, M.: Matching biodiversity and ecology ontologies: challenges and evaluation results. The Knowledge Engineering Review  \textbf{35}, ~e9 (2020)

\bibitem{karpathy_knn_svm}
Karpathy, A.: knn vs svm. \url{https://github.com/karpathy/randomfun/blob/master/knn_vs_svm.ipynb} (Accessed: 2024-12-17)

\bibitem{rag}
Lewis, P., Perez, E., Piktus, A., Petroni, F., Karpukhin, V., Goyal, N., K\"{u}ttler, H., Lewis, M., Yih, W.t., Rockt\"{a}schel, T., Riedel, S., Kiela, D.: Retrieval-augmented generation for knowledge-intensive nlp tasks. In: Proceedings of the 34th International Conference on Neural Information Processing Systems. NIPS '20, Curran Associates Inc., Red Hook, NY, USA (2020)

\bibitem{liu2024context}
Liu, S., Ye, H., Xing, L., Zou, J.: In-context vectors: making in context learning more effective and controllable through latent space steering. In: Proceedings of the 41st International Conference on Machine Learning. pp. 32287--32307 (2024)

\bibitem{mohammadi2020evaluating}
Mohammadi, M., Rezaei, J.: Evaluating and comparing ontology alignment systems: an mcdm approach. Journal of Web Semantics  \textbf{64},  100592 (2020)

\bibitem{nas2023mse}
Nas, E., Huschka, M.: Mse benchmark. \url{https://github.com/EngyNasr/MSE-Benchmark} (2023)

\bibitem{nasr2020evaluation}
Nasr, E.: Evaluation of Automatic Ontology Matching for Materials Sciences and Engineering. Ph.D. thesis, Master’s thesis, Albert Ludwig University of Freiburg, Germany (2020)

\bibitem{oaei2024}
{OAEI Community}: Ontology alignment evaluation initiative (oaei) (2024), \url{https://oaei.ontologymatching.org}

\bibitem{okgreece_alignment}
okgreece: Alignment: A collaborative, system-aided ontology matching application. \url{https://github.com/okgreece/Alignment} (2024), accessed: 2024-12-02

\bibitem{portisch2022background}
Portisch, J., Hladik, M., Paulheim, H.: Background knowledge in ontology matching: A survey. Semantic Web pp. 1--55 (2022)

\bibitem{qiang2023agent}
Qiang, Z., Wang, W., Taylor, K.: Agent-om: Leveraging llm agents for ontology matching. arXiv preprint arXiv:2312.00326  (2023)

\bibitem{sbert}
Reimers, N., Gurevych, I.: Sentence-bert: Sentence embeddings using siamese bert-networks. In: Proceedings of the 2019 Conference on Empirical Methods in Natural Language Processing. Association for Computational Linguistics (11 2019), \url{https://arxiv.org/abs/1908.10084}

\bibitem{ibm2024ontologyalignment}
Research, I.: Ontology alignment (2024), \url{https://github.com/IBM/ontology-alignment}, accessed: December 2, 2024

\bibitem{fuzzywuzzy}
Richard, A.: fuzzywuzzy. \url{https://pypi.org/project/fuzzywuzzy/} (2024)

\bibitem{Rupp.2021}
Rupp, C., {The SOPHISTs}: Requirements-Engineering UND -Management: Das Handbuch für Anforderungen in jeder Situation. Hanser (2021)

\bibitem{shalagin2024survey}
Shalagin, N.: A survey on natural language semantic search algorithms. International Journal of Open Information Technologies  \textbf{12}(9),  11--21 (2024)

\bibitem{OAEI23}
Shvaiko, P., Euzenat, J., Jim{\'{e}}nez{-}Ruiz, E., Hassanzadeh, O., Trojahn, C. (eds.): Proceedings of the 18th International Workshop on Ontology Matching co-located with the 22nd International Semantic Web Conference {(ISWC} 2023), Athens, Greece, November 7, 2023, {CEUR} Workshop Proceedings, vol.~3591. CEUR-WS.org (2023)

\bibitem{silva2024complex}
Silva, M., Faria, D., Pesquita, C.: Complex multi-ontology alignment through geometric operations on language embeddings. In: ECAI. pp. 1333--1340 (2024)

\bibitem{sousa2025complex}
Sousa, G., Lima, R., Trojahn, C.: Complex ontology matching with large language model embeddings. arXiv preprint arXiv:2502.13619  (2025)

\bibitem{taboada2025ontology}
Taboada, M., Martinez, D., Arideh, M., Mosquera, R.: Ontology matching with large language models and prioritized depth-first search. arXiv preprint arXiv:2501.11441  (2025)

\bibitem{TheSophists.2016}
{The SOPHISTs}: Requirements Engineering - A Short RE Primer (2016)

\bibitem{touvron2023llama2openfoundation}
Touvron, H., Martin, L., Stone, K., Albert, P., Almahairi, A., Babaei, Y., Bashlykov, N., Batra, S., Bhargava, P., Bhosale, S., Bikel, D., Blecher, L., Ferrer, C.C., Chen, M., Cucurull, G., Esiobu, D., Fernandes, J., Fu, J., Fu, W., Fuller, B., Gao, C., Goswami, V., Goyal, N., Hartshorn, A., Hosseini, S., Hou, R., Inan, H., Kardas, M., Kerkez, V., Khabsa, M., Kloumann, I., Korenev, A., Koura, P.S., Lachaux, M.A., Lavril, T., Lee, J., Liskovich, D., Lu, Y., Mao, Y., Martinet, X., Mihaylov, T., Mishra, P., Molybog, I., Nie, Y., Poulton, A., Reizenstein, J., Rungta, R., Saladi, K., Schelten, A., Silva, R., Smith, E.M., Subramanian, R., Tan, X.E., Tang, B., Taylor, R., Williams, A., Kuan, J.X., Xu, P., Yan, Z., Zarov, I., Zhang, Y., Fan, A., Kambadur, M., Narang, S., Rodriguez, A., Stojnic, R., Edunov, S., Scialom, T.: Llama 2: Open foundation and fine-tuned chat models (2023), \url{https://arxiv.org/abs/2307.09288}

\bibitem{zhang2024large}
Zhang, S., Dong, Y., Zhang, Y., Payne, T.R., Zhang, J.: Large language model assissted multi-agent dialogue for ontology alignment. In: Proceedings of the 23rd International Conference on Autonomous Agents and Multiagent Systems. pp. 2594--2596 (2024)

\end{thebibliography}


\begin{thebibliography}{8}
\bibitem{ref_article1}
Author, F.: Article title. Journal \textbf{2}(5), 99--110 (2016)

\bibitem{ref_lncs1}
Author, F., Author, S.: Title of a proceedings paper. In: Editor,
F., Editor, S. (eds.) CONFERENCE 2016, LNCS, vol. 9999, pp. 1--13.
Springer, Heidelberg (2016). \doi{10.10007/1234567890}

\bibitem{ref_book1}
Author, F., Author, S., Author, T.: Book title. 2nd edn. Publisher,
Location (1999)

\bibitem{ref_proc1}
Author, A.-B.: Contribution title. In: 9th International Proceedings
on Proceedings, pp. 1--2. Publisher, Location (2010)

\bibitem{ref_url1}
LNCS Homepage, \url{http://www.springer.com/lncs}, last accessed 2023/10/25
\end{thebibliography}
\end{document}